\documentclass[11pt]{article}

\usepackage[final]{acl}
\usepackage{times}
\usepackage{booktabs}
\usepackage[russian,english]{babel}
\usepackage{url}
\usepackage[T1]{fontenc}
\usepackage[utf8]{inputenc}
\usepackage{microtype}
\usepackage{inconsolata}
\usepackage{graphicx}
\usepackage[table]{xcolor}

\definecolor{LayerA}{HTML}{4292C6}
\definecolor{LayerB}{HTML}{6BAED6}
\definecolor{LayerC}{HTML}{9ECAE1}
\definecolor{LayerD}{HTML}{C6DBEF}
\definecolor{LayerE}{HTML}{DEEBF7}

\newcommand{\ph}[1]{\textlangle\textit{#1}\textrangle}

\title{Script Choice in LLMs: Evidence for Late-Layer Commitment}

\author{David Kletz\footnotemark[2], Sandra Mitrović\footnotemark[2], \hspace{0.8mm}   {\bf Itay Sabato\footnotemark[4],} \hspace{0.8mm} {\bf  Ljiljana Dolamić\footnotemark[3],} \hspace{0.8mm} {\bf  Fabio Rinaldi\footnotemark[2]}\\
\footnotemark[2] SUPSI, IDSIA, Switzerland\\
\footnotemark[3] armasuisse, Science \& Technology, Switzerland\\
\footnotemark[4] Independent Researcher\\
{\tt \{david.kletz, sandra.mitrovic, fabio.rinaldi\}@supsi.ch} \\
{\tt ljiljana.dolamic@armasuisse.ch}\\
{\tt itaysabato@gmail.com}\\}

\begin{document}
\maketitle
\begin{abstract}
In this paper, we investigate how script knowledge is distributed across the layers of LLMs using two complementary interpretability methods: logistic regression probing and logit-lens analysis. Our probing experiments reveal a clear asymmetry: both the input script and the instructed output script are encoded in the earliest layers of the network, while, in contrast, commitment to the actual output script emerges only in the final layers, with the model's intermediate representations defaulting to Latin throughout most of the layers. This two-stage process is confirmed by logit-lens analyses, which show that script commitment consistently occurs at the very last layers of the LLMs. Together with the weaker script-following performance observed in smaller models, these results form a converging body of evidence linking script commitment to model depth, with broader implications for the design of sufficiently deep, inclusive multilingual architectures.
\end{abstract}

\section{Introduction}

Large Language Models (LLMs) can generate text across scripts, from Latin and Cyrillic to Arabic and Devanagari \cite{costa2022no, workshop2022bloom, apertus2025apertus}. This ability appears simple behaviorally, but internally, the model must first identify the script of the input, infer the script required for the output, and ultimately shift its probability mass toward tokens in the appropriate script.
Transformer models process different aspects of linguistic knowledge at different depths \cite{tenney2019bert, peters2018dissecting, belinkov2022probing}, as a consequence, these sub-tasks might not be executed simultaneously at a single point in the network, but instead distributed across layers.

This distributional assumption motivates our central research question: where, across the layers of an LLM, does script knowledge emerge, and does the answer differ depending on whether we consider the recognition of the input script, the planning of the output script, or the commitment of the output distribution to a target writing system? Prior work supports the plausibility of such layer-wise specialization beyond syntax and semantics. At the behavioral level, \citet{liang-etal-2024-abseval} confirm that LLMs possess a degree of script planning capability, suggesting that output script selection is a deliberate, learnable operation rather than a mere surface artefact. At a coarser granularity, \citet{pochinkov2024extractingparagraphsllmtoken, pochinkov2025parascopeslanguagemodelsactivations} show that commitment to paragraph-level content may be a general property of autoregressive generation, which we expect to extend to script-level commitment as well.
Evidence further suggests that adjacent layers may not always build cooperatively on one another \cite{patrawala2026llm}, that simpler tasks require fewer layers \cite{fan2025not}, and that high-resource language knowledge tends to be consolidated in deeper layers \cite{li2025exploring}.
\nocite{tamo2026linguamap}

Script processing, however, has not yet been studied from this perspective, even though this is a particularly interesting case: it intervenes at multiple stages of generation, from input recognition to output realization. Most closely related to our work is RomanLens~\cite{saji-etal-2025-romanlens}, which pairs the logit lens with activation patching to study the mechanism of romanization and language transfer, and reports a related finding: intermediate layers often represent non-Roman-script targets in Romanized form before the model transitions to native script. Our work differs both methodologically, pairing the logit lens with linear probing rather than patching, and conceptually, distinguishing input script, instructed output script, and the script actually produced.

In this paper, separation of these three scripts is central and we investigate how script knowledge is distributed across the layers of large language models. We focus on the Qwen 2.5 Instruct family (0.5B–32B), which provides a controlled experimental setting.

To trace script processing through the model's layers, we employ two methods. In Section~\ref{sec:probing-results}, we train probes \cite{alain2016understanding} on the hidden state of the final prompt token at every layer to identify where script information becomes linearly accessible, distinguishing input script, instructed output script, script produced by the LLM. Then, in Section~\ref{sec:logitlens-results}, we apply a logit-lens analysis \cite{lesswrongInterpretingGPT} to track when the output distribution commits its output distribution to the target script. Our experiments cover nine scripts\footnote{Data is available on our GitHub repo \url{https://github.com/IDSIA-NLP/LLMScriptCommitment}}: Latin, Cyrillic, Arabic, Hebrew, Devanagari, Korean, Japanese, Chinese, and Armenian.
Finally, showing that commitment to non-Latin scripts emerges only in the last layers of the LLM, in Section~\ref{s:discussion} we discuss this pattern as a potential, correlational explanation for the weaker performance of smaller models on non-Latin scripts.

\section{Experimental Setup}\label{sec:setup}

\subsection{Models}
We work with the Qwen~2.5 Instruct family across six sizes: 0.5B, 1.5B,
3B, 7B, 14B, and 32B\footnote{The 14B and 32B checkpoints were run under 4-bit quantization}. All checkpoints share the same pre-training corpus
\cite{qwen2025qwen25technicalreport}, so behavioral differences across scales reflect
architecture rather than knowledge. The models span 24 to 64 transformer
layers, making this a good suite for studying the effect of depth. Note however that depth increases non-monotonically with parameter count: the 3B model (36 layers) is deeper than the 7B (28 layers).
The full architecture table and quantization details are in Appendix~\ref{app:models}.

\subsection{Data and Prompts}
Prompts are drawn from nine source languages: English, Russian, Arabic,
Chinese, Japanese, Korean, Hindi, Armenian, and Hebrew. Each language
appears in two conditions: a canonical condition, where the model is instructed to produce output in the same script as the input, and a cross-script condition, where it is asked to use a different script. For all languages except English, the cross-script target is Latin. This yields 17 language-to-script mappings in total.

Within each mapping, prompts are drawn from three families: sentences, words, and questions. Each prompt is instantiated across four instruction variants that range from explicit (\textit{``write the following using only Cyrillic characters''}) to concise (\textit{``answer in Cyrillic''}). English and Russian are represented by 274 input items each (80 sentences, 114 words, 80 questions); the remaining seven languages by 70 items each (20/30/20).
The full dataset comprises 8,304 prompts. Each model sees the
same set. Detailed prompt templates are listed in Appendix~\ref{app:prompts}.

\subsection{Script Classification}
\label{sec:script}
The script of a token is determined by the Unicode block name of its first character. We define ten script classes: one for each available input script and a ``Neutral'' class for tokens whose first character is any character that doesn't have a script. Tokens in the Neutral class are excluded from script-specific analysis. This classifier is applied at two points: to the prompt, to derive the input-script label, and to the first generated token\footnote{We focus on the first generated token because it directly reflects the initial script-selection decision before auto-regressive feedback effects} of each model response, to derive the used-output-script label.

\subsection{Probing}
\label{sec:probing}
For each (model, prompt) pair, we record the hidden state at the
last prompt token at every layer (the embedding output from layer~0 and each of the $L-1$ transformer block outputs) extracted during the prefill step of the generation forward pass, without a second inference. We train one independent logistic regression probe per (target, layer) pair against three targets: \textit{Input script (9 classes)}: the source script of the prompt, \textit{Requested-output script (9 classes)}: the script the prompt instructs the model to use, and \textit{Used-output script} (10 classes): the script the model actually produced, including a Neutral/Other bucket for non-classifiable outputs.

Probes are a logistic regression with $\ell_2$ regularization trained on per-feature standardized activations. High accuracy from such a probe indicates that the target is geometrically accessible (hidden states of different classes are linearly separable in activation space, \citealp{hewitt-liang-2019-designing}) in the representation at that layer, not merely encoded in some nonlinear fashion. We use a 70/30 stratified train/test split, averaged across five independent random splits to reduce variance. Our primary metric is balanced accuracy (mean per-class recall), which is insensitive to the class imbalance introduced by the Neutral bucket in the used-output target; we also report AUC and raw accuracy as secondary metrics. Full hyperparameters are in Appendix~\ref{app:probes}.

\subsection{Logit Lens}
\label{sec:logitlens}
The logit lens \cite{lesswrongInterpretingGPT} projects each intermediate representation to obtain a distribution over the full vocabulary at every layer. We apply this to the same last-prompt-token hidden states used in section~\ref{sec:probing}, requiring no additional forward pass. At each layer~$\ell$, we extract the top-1 token of the projected distribution and classify its script using the classifier of section~\ref{sec:script}. We define the script commitment layer as the earliest~$\ell$ at which the top-1 token belongs to the target script\footnote{Examples without such layer are considered as failures}. We report two quantities: (i)~the fraction of examples where the model's generated token belongs to the correct script, and (ii)~the mean script commitment layer over successful examples, broken down by target script and model size.

\section{Encoding of Script}
\label{sec:probing-results}

\begin{figure}
    \centering
    \includegraphics[width=1.\linewidth]{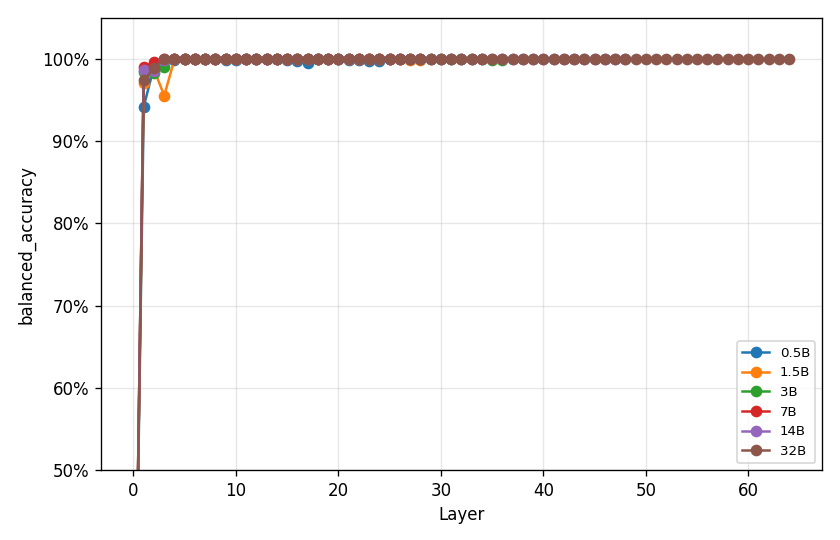}
    \caption{Balanced accuracy of the requested-output script probe across layers, for all six Qwen~2.5 model sizes.}
    \label{fig:probe-requested}
\end{figure}

\begin{figure}
    \centering
    \includegraphics[width=1.\linewidth]{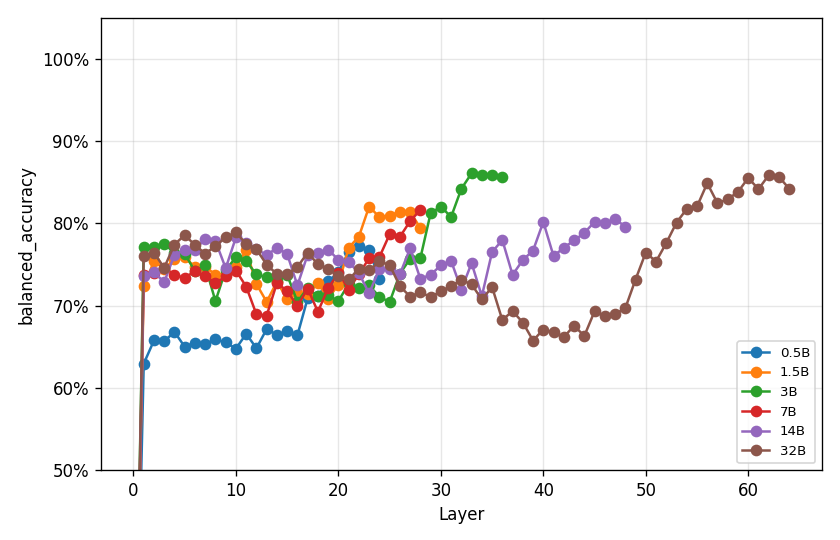}
    \caption{Balanced accuracy of the used-output script probe across layers.}
    \label{fig:probe-used}
\end{figure}

At layer~0, all targets are at chance for every model (${\simeq}\,0.11$ for the 9-class targets and ${\simeq}\,0.09$ for used-output), confirming that token embeddings carry no linearly accessible script information. This, in fact, is expected and serves as a sanity check, since all our prompts end with ``Answer:'' (see Appendix~\ref{app:prompts}) and, hence, have as the last prompt token always ``:''. However, at layer~1, input script already reaches ${\simeq}\,0.98$ balanced accuracy across all model sizes.

Figure~\ref{fig:probe-requested} shows balanced accuracy for the requested-output target. All six models reach 1.00 by layer 3--5, regardless of size or depth: the script the model was instructed to use is fully encoded within the first ${\sim}5$--$20$\% of the network. Notably, balanced accuracy decreases slightly in the final layers of each model suggesting that the explicit representation of the requested script is gradually displaced as the network shifts from encoding the instruction to computing the output distribution.

In Figure~\ref{fig:probe-used}, the used-output target rises more slowly, peaks at 82--100\% of model depth, and never saturates, with balanced accuracy ranging from 0.77 (0.5B) to 0.86 (3B) at peak. The maximum is set in part by each model's actual script-following rate (0.64--0.83 strict, across sizes). Together, the two figures establish the core dissociation: the model encodes what it should produce within the first few layers, but what it will actually produce only becomes readable in the final layers.
Section~\ref{sec:logitlens-results} examines what happens in between.

\section{Late Commitment to Output Script}
\label{sec:logitlens-results}

\begin{figure*}
    \centering
    \includegraphics[width=1.\linewidth]{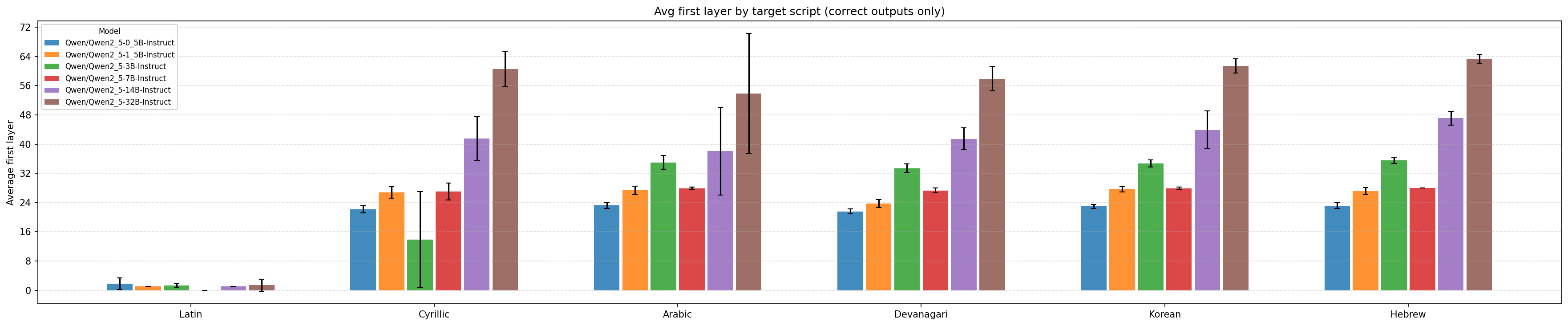}
    \caption{Average first layer of script commitment by target script and model size. Error bars show standard deviation across prompts. Hollow profiles show the maximum each model can reach (number of layers).}
    \label{fig:commitment-layer}
\end{figure*}

Figure~\ref{fig:commitment-layer} shows the mean script commitment layer per target script and model size, computed over successful generations only. It shows an important contrast between Latin and all other scripts: the top-1 predicted token at the generation position commits to Latin within the first 1--2 layers across all six models regardless of the source language of the prompt. Non-Latin script generation requires a late override: for Cyrillic, Arabic, Devanagari, Korean, and Hebrew, the commitment layer consistently falls above $85\%$ of model depth, often in the final three to five layers.

The commitment layer grows with model depth in absolute terms: for Cyrillic, it rises from ${\sim}$22 layers (0.5B, 24 total) to ${\sim}$60 layers (32B, 64 total), while remaining at roughly the same relative position across most models and scripts. One notable exception is Cyrillic for the 3B model, which commits at layer ${\sim}$14 out of 36.
Armenian, Chinese, and Japanese yield too low success rates across all model sizes and are absent from Figure~\ref{fig:commitment-layer}. Detailed results are in Appendix~\ref{app:detailed_resuls}, and show that when their output is the correct one, they exhibit the same late-commitment pattern as the other non-Latin scripts.

\begin{table}[ht]
\centering
\begin{tabular}{ll}
\toprule
\textbf{Layer} & \textbf{Prediction} \\
\midrule
\rowcolor{LayerA}\textbf{h\_out}        & \textbf{\texttt{\foreignlanguage{russian}{' К'}}}\\
\rowcolor{LayerB}\texttt{h\textsubscript{h34}\_out} & \textbf{\texttt{\foreignlanguage{russian}{' К'}}} \\
\rowcolor{LayerC}\texttt{h\textsubscript{h33}\_out} & \texttt{'-capital'} \\
\rowcolor{LayerD}\texttt{h\textsubscript{h32}\_out} & \texttt{'-capital'} \\
\rowcolor{LayerE}\texttt{h\textsubscript{h31}\_out} & \texttt{'-capital'} \\
\bottomrule
\end{tabular}
\caption{Logit-lens top-1 prediction at the generated token position across the last 5 layers of \textsc{Qwen2.5-3B-Instruct} when asked to write ``Capital'' using the Cyrillic script. The \foreignlanguage{russian}{' К'} character is a Cyrillic character.}
\label{tab:heatmap}
\end{table}

Table~\ref{tab:heatmap} illustrates the underlying mechanism.
The heatmap shows the top-1 logit-lens token at each layer and
each prompt position for a representative English-to-Cyrillic
generation (target: \textit{\foreignlanguage{russian}{``Капитал''}}). Through most of the
network, the top-1 token at the generation position is the
semantically correct Latin token (\textit{``capital''}). Only in the final five to eight layers does the prediction switch to the Cyrillic equivalent. The Cyrillic token is not arbitrary: it is the transliteration of the Latin prediction held in the preceding layers. This two-stage process accounts for both the late commitment observed in Figure~\ref{fig:commitment-layer} and suggests that the failure of models to complete it might be due to the insufficient model depth. Additionally, this finding corroborates the one of RomanLens~\cite{saji-etal-2025-romanlens}, that independently reports a similar late-layer script transition on different models.

\section{Models' Depth Bottleneck Hypothesis and Smaller Models Failures}\label{s:discussion}

\begin{figure}[h]
    \centering
    \includegraphics[width=1.\linewidth]{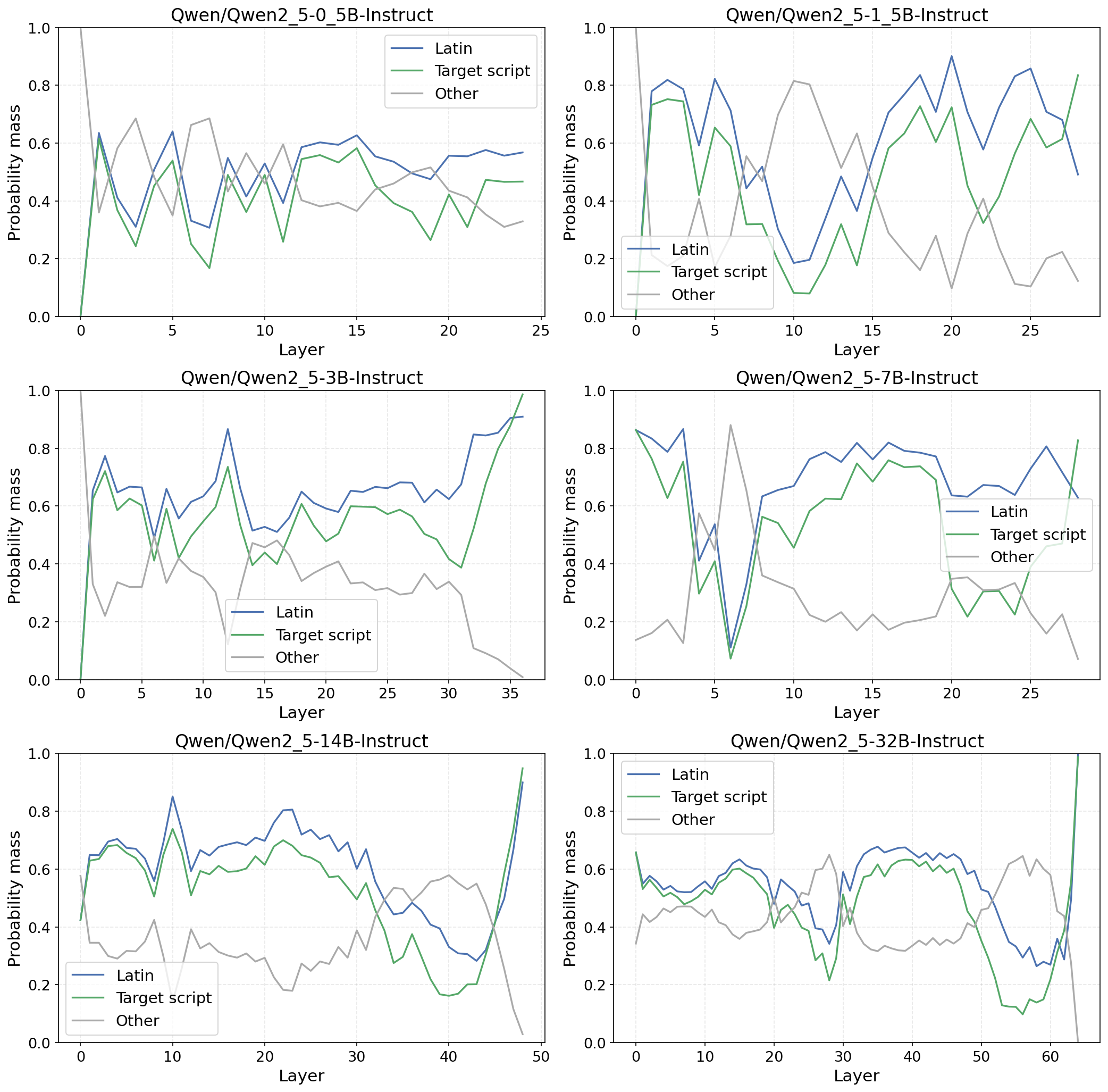}
    \caption{Probability mass assigned to tokens of Latin, target and any other script across layers.}
    \label{fig:probmass}
\end{figure}

Our results show that script encoding and identification are distributed across the model's layers. Input script identity is encoded at the first transformer block (layer~1, ${\sim}98\%$ balanced accuracy). The instructed output script is also captured and encoded within the first five layers (${\sim}5$--$20\%$ of depth). The vocabulary distribution, however, does not commit to non-Latin scripts until the final layers (${\geq}85\%$ of depth), and does so imperfectly in proportion to each model's depth. More specifically, Figure~\ref{fig:probmass} presents the total probability mass assigned to Latin tokens,  tokens in the requested target script, and tokens in any other script, at each layer. It shows that for most of each model's depth, Latin mass matches or exceeds target-script mass, with target-script mass surpassing Latin mass only toward the final layers. Qwen2.5-0.5B is the one exception, where Latin mass stays at or above target-script mass even at the very last layer, consistent with the observed model lack of depth.

\paragraph{Linear accessibility does not imply functional use}
The gap between layers of target script encoding and layers of realization in the distribution reflects a distinction between encoding information and using it. The model keeps the target script information accessible from layer~5 onward, yet its output distribution does not reflect this. We interpret the intervening computation as the network working through the semantic content of the answer in its default register (Latin), with script conversion deferred to the final layers. The slight decrease in requested-output probe accuracy observed at the end of the network (Section~\ref{sec:probing-results}) is consistent with this view: as the final layers shift from maintaining the instruction to executing it, the explicit linear representation of the target script is partially displaced by the output distribution it is generating.

The non-monotonic relation between model size in parameters and script-following performance shows that depth is the important parameter for this task. Non-Latin generation requires enough layers to execute the late-stage script conversion that our logit-lens analysis makes visible. A model with too few layers defaults to a neutral or Latin output, regardless of how well it has encoded the instruction.

We thus interpret the smaller models’ struggle with non-Latin scripts as supportive of the hypothesis that limited model depth may constitute a bottleneck to completing the late-stage script conversion process, rather than reflecting a failure of knowledge or representation.

\section{Conclusion}

In this paper, we have shown that script processing in LLMs is not a monolithic operation. Rather, input identification, instruction encoding, and output script commitment emerge at different stages of the network. The output vocabulary distribution is modified only after the model has formed an internal representation of what it intends to produce, and the gap between representation and script realization is resolved only in the final layers.

The observed pattern offers a potential mechanistic explanation for the weaker script-following performance of smaller models: rather than failing to encode the target script instruction, it suggests that these models lack layers to complete the late conversion process.

As such, our findings have broader implications for the development of multilingual LLMs: improving support for underrepresented writing systems may require not only better representations, but also sufficiently deep architectures to translate those representations into reliable generation.

Future work should also investigate how LLMs acquire the ability to transliterate their top Latin prediction into the target script, and complement our script-based analysis with a breakdown by language family.

\section*{Limitations}

We acknowledge the following limitations.
First, all experiments are conducted on a single model family, Qwen 2.5 Instruct. It remains an open question whether the observed patterns generalize to other architectures.
Second, the two largest models (14B and 32B) were run under 4-bit quantization due to hardware constraints.
Third, Armenian, Chinese, and Japanese exhibited too low script-following success rates across all model sizes to be included in the main logit-lens analysis.

Finally, our probing and logit-lens analyzes both rely on a single token position: the last token of the prompt, which serves as the prediction site for the first generated token.

\section*{Acknowledgments}
The work described in this paper has been partially funded by the ``\textit{Language Preference and Hallucination Investigation in Multilingual RAG (LPHI-mRAG)}'' project, funded by armasuisse S\&T, Switzerland.

\bibliography{custom}

\appendix

\section{Models}\label{app:models}
See details in Table~\ref{tab:models}.
Large models (Qwen2.5-14B and Qwen2.5-32B) were loaded with 4-bit) quantization via BitsAndBytes, using double quantization and bfloat16 compute.

\begin{table}[h]
\centering
\resizebox{\columnwidth}{!}{%
\begin{tabular}{lcccc}
\toprule
Model & Params & Layers & Hidden dim & Attn / KV heads \\
\midrule
Qwen2.5-0.5B-Instruct & $\sim$0.5B & 24 &  896 & 14 / 2 \\
Qwen2.5-1.5B-Instruct & $\sim$1.5B & 28 & 1536 & 12 / 2 \\
Qwen2.5-3B-Instruct   & $\sim$3B   & 36 & 2048 & 16 / 2 \\
Qwen2.5-7B-Instruct   & $\sim$7B   & 28 & 3584 & 28 / 4 \\
Qwen2.5-14B-Instruct  & $\sim$14B  & 48 & 5120 & 40 / 8 \\
Qwen2.5-32B-Instruct  & $\sim$32B  & 64 & 5120 & 40 / 8 \\
\bottomrule
\end{tabular}}
\caption{Qwen2.5 models used.}
\label{tab:models}
\end{table}

\section{Prompts}\label{app:prompts}

All prompts end by asking the answer with the word ``\texttt{Answer:}'' so that the
model's first generated token is the start of the answer. Placeholders
are shown in angle brackets: \ph{target\_script} is the name of the
requested script (e.g.\ \textit{Latin}, \textit{Cyrillic}), and
\ph{input} is the source sentence, word, or question.

\subsection*{Sentences}

\begin{tabular}{@{}lp{0.6\linewidth}@{}}
\toprule
Variant & Template \\
\midrule
\texttt{write\_using} &
    Write the following sentence using the \ph{target\_script} script:
    ``\ph{sentence}'' \\[4pt]
\texttt{rewrite} &
    Rewrite this sentence in \ph{target\_script}: ``\ph{sentence}'' \\[4pt]
\texttt{write\_chars} &
    Write the sentence ``\ph{sentence}'' using \ph{target\_script}
    characters. \\[4pt]
\texttt{how\_write} &
    How would you write the following sentence in \ph{target\_script}?
    ``\ph{sentence}'' \\
\bottomrule
\end{tabular}

\subsection*{Words}

\begin{tabular}{@{}lp{0.6\linewidth}@{}}
\toprule
Variant & Template \\
\midrule
\texttt{write\_using} &
    Write the word ``\ph{word}'' using the \ph{target\_script} script. \\[4pt]
\texttt{write\_in} &
    Write the following word in \ph{target\_script}: \ph{word} \\[4pt]
\texttt{how\_write} &
    How do you write ``\ph{word}'' in the \ph{target\_script} script? \\[4pt]
\texttt{spell} &
    Spell the word ``\ph{word}'' using \ph{target\_script} characters. \\
\bottomrule
\end{tabular}

\subsection*{Questions}

\begin{tabular}{@{}lp{0.6\linewidth}@{}}
\toprule
Variant & Template \\
\midrule
\texttt{answer\_using} &
    Answer the following question using the \ph{target\_script} script:
    \ph{question} \\[4pt]
\texttt{inline} &
    \ph{question} (Answer using the \ph{target\_script} script.) \\[4pt]
\texttt{terse} &
    Answer in \ph{target\_script}: \ph{question} \\[4pt]
\texttt{only\_chars} &
    Respond to the following question using only \ph{target\_script}
    characters: \ph{question} \\
\bottomrule
\end{tabular}

\section{Probes}\label{app:probes}
$\ell_2$ regularisation ($C\!=\!1$), and L-BFGS optimisation, on a
single 70/30 train/test split of 1,000 prompts after per-feature
standardisation.

\section{Script-Following Behavior}
\label{app:instr_following}

We check each model's behavioral compliance
with the requested script.  The strict follow rate is the fraction of
prompts for which the used-output script matches the requested script; the
lax follow rate additionally counts ``Other'' and ``Neutral'' outputs as
successes
(Table~\ref{tab:follow}).

\begin{table}[h]
\centering
\small
\begin{tabular}{lcc}
\toprule
Model & Strict & Lax \\
\midrule
Qwen2.5-0.5B-Instruct & 0.641 & 0.797 \\
Qwen2.5-1.5B-Instruct & 0.719 & 0.846\\
Qwen2.5-3B-Instruct   & 0.831 & 0.920 \\
Qwen2.5-7B-Instruct   & 0.799 & 0.898 \\
Qwen2.5-14B-Instruct  & 0.766 & 0.838 \\
Qwen2.5-32B-Instruct  & 0.754 & 0.833\\

\bottomrule
\end{tabular}
\caption{Script-following rates.}
\label{tab:follow}
\end{table}

\section{Input language}\label{app:input_lang}
Input language is the easiest target by a wide margin: it is fully determined
by the surface tokens of the prompt, requiring no cross-token integration.  At
layer~0 (embedding output) all probes sit at chance (balanced accuracy
0.47--0.54); a single transformer block suffices to drive every metric to
$\approx$1.0 in every model.

\begin{figure}
    \centering
    \includegraphics[width=1.\linewidth]{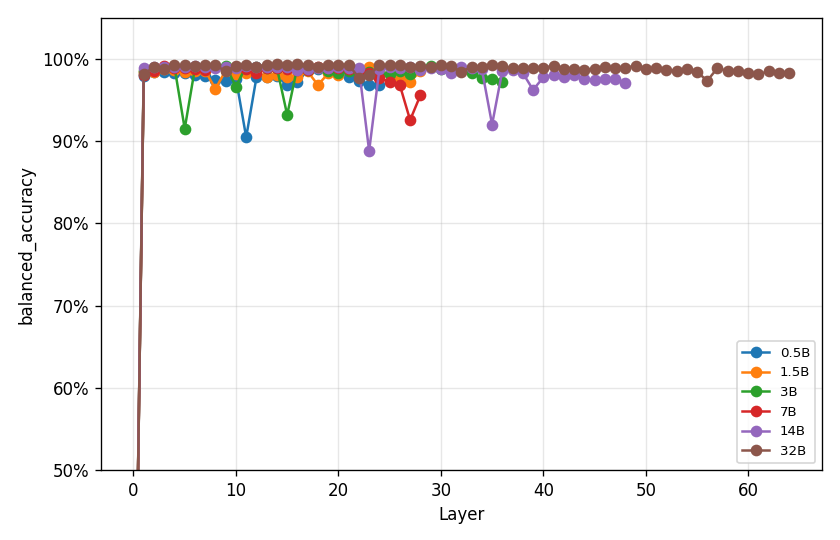}
    \caption{Balanced accuracy of the input-language probe across layers, for all six Qwen models.}
    \label{fig:probe-input}
\end{figure}

\section{Detailed Results}\label{app:detailed_resuls}

\begin{figure*}
    \centering
    \includegraphics[width=1.15\linewidth]{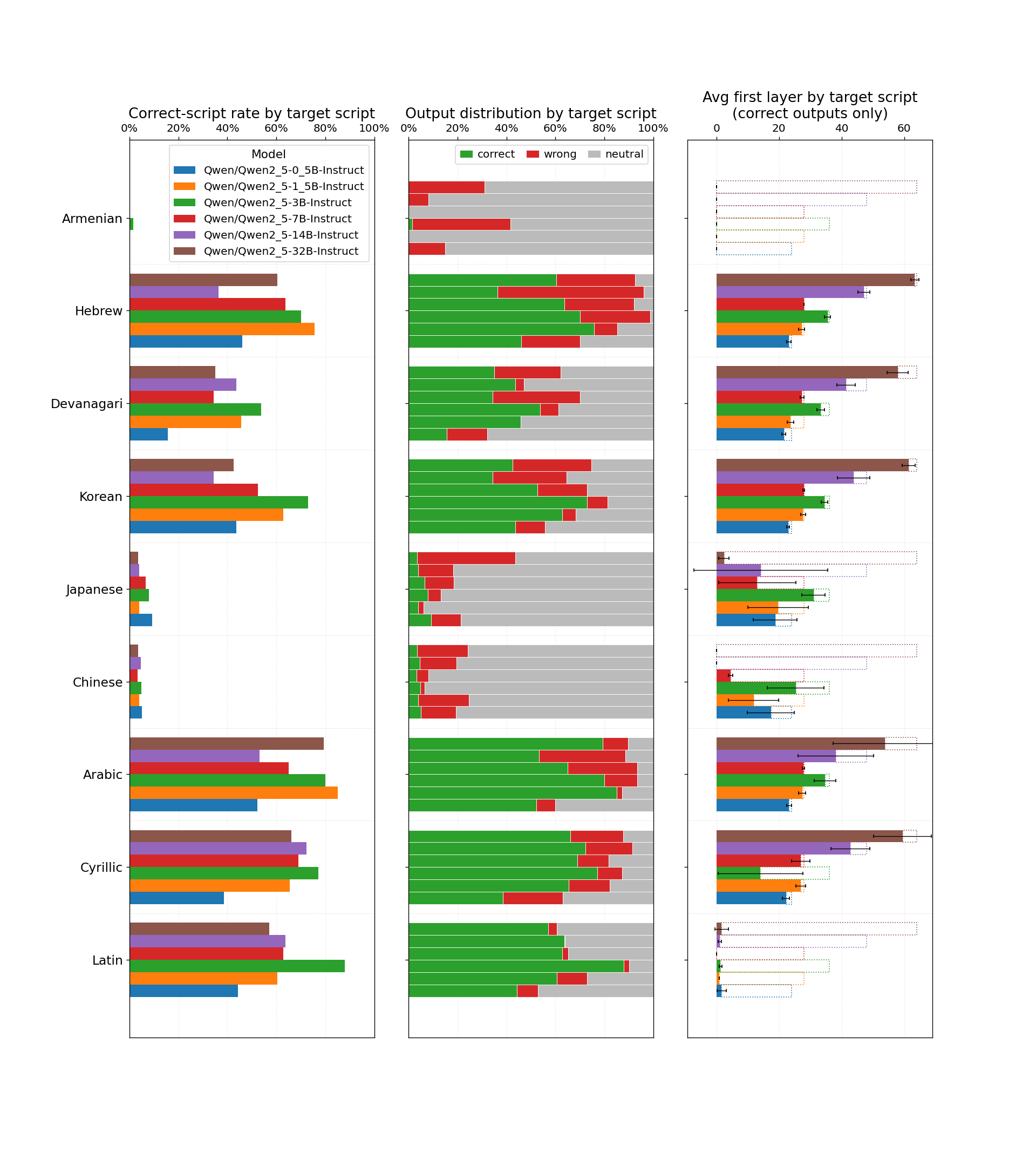}
    \caption{Per-script breakdown across all six Qwen models. Left: correct-script generation rate by target script. Center: output distribution by target script, broken down into correct, wrong, and neutral outputs. Right: average first commitment layer by target script for correct outputs only.}
    \label{fig:detailed_results}
\end{figure*}

Figure~\ref{fig:detailed_results} illustrates per-script breakdown across all six Qwen models.

\begin{figure*}[t]
    \centering
    \includegraphics[width=0.72\linewidth]{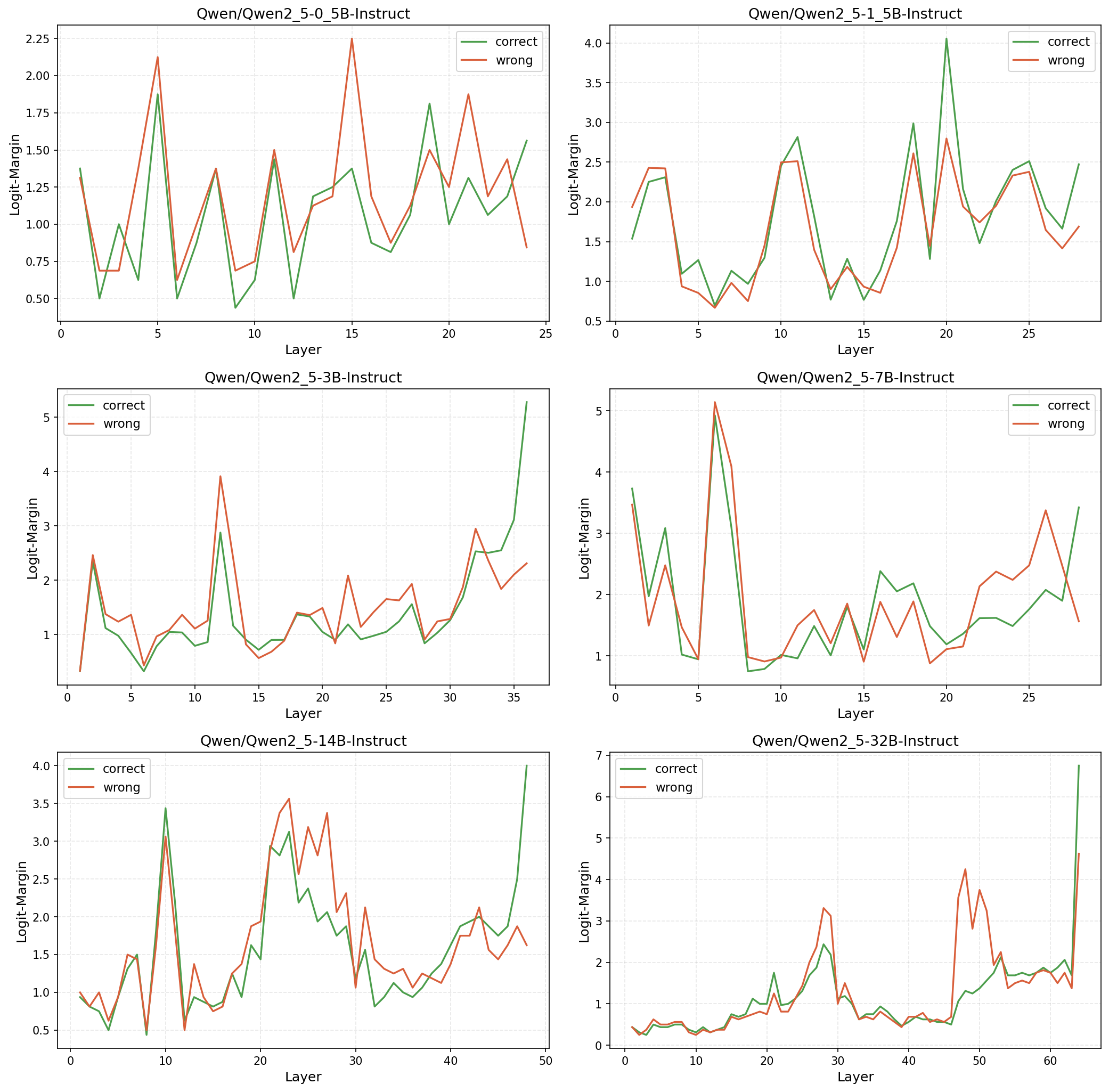}
    \caption{Logit margin between the top-1 token and the highest-ranked token from different script, across layers.}
    \label{fig:margin}
\end{figure*}

To investigate the distribution of probability mass beyond the top-1 token at each layer, we calculate the logit margin between the top-1 token and the highest-ranked token belonging to a different script (Figure~\ref{fig:margin}).
Across all six models, the logit margin stays modest (from 0.5 to 5), before increasing in the final layers, with the two models 14B and 32B showing the largest increase in margin at the final layer. This confirms top-1 at the commitment layer is not a weak tie-break. At the last layer, prompts whose output matches the requested script show a larger margin than those that do not.

\section{Wrong Scripts Generation}\label{app:wrong_script}

\begin{figure}
    \centering
    \includegraphics[width=1.\linewidth]{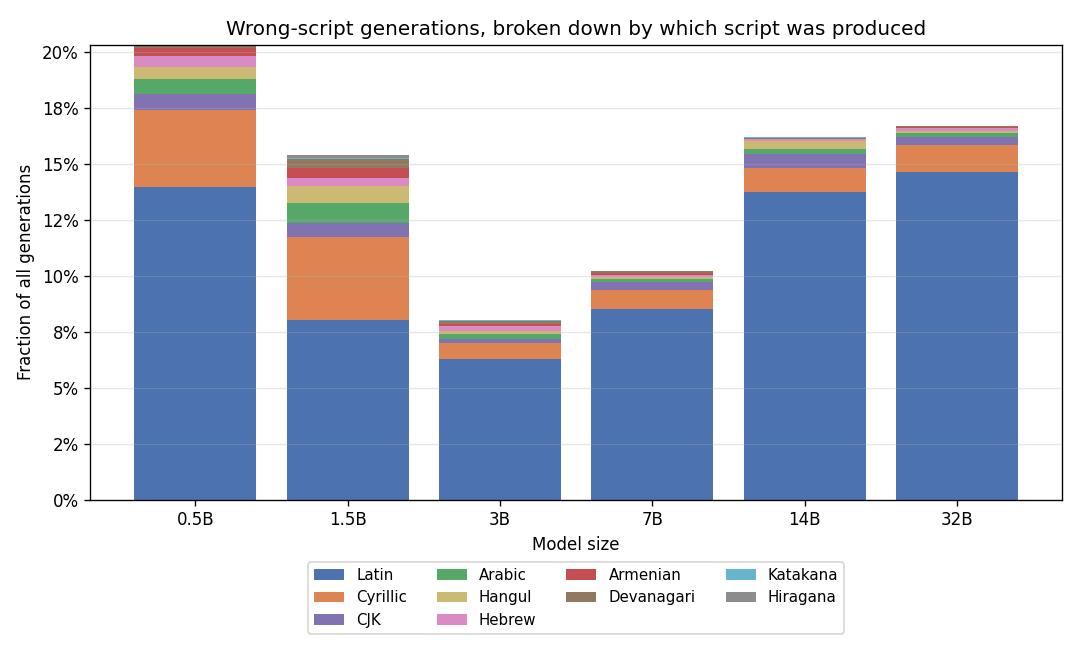}
\caption{Breakdown of wrong-script generations by the script actually
produced, per model.}
    \label{fig:wrong_script_bd}
\end{figure}

For each model, we add in Figure~\ref{fig:wrong_script_bd}
a breakdown of the prompts whose used-output script does not match the
requested script, by the script that was actually produced.

\section{Computational Details}
All experiments were run on an NVIDIA L40S GPU (46\,GB VRAM).
All models were run with greedy decoding (\texttt{do\_sample=False}) and a maximum budget of 256 new tokens. Generation was stopped at the first token whose decoded text contained a classifiable Unicode character.

Probes were trained on GPU using a batched logistic regression implemented in PyTorch, fitting all layers of a given target simultaneously in a single Adam optimization loop (learning rate 0.1, up to 1000 epochs).
\end{document}